\documentclass[11pt, a4paper, logo, onecolumn, copyright, numbering]{googledeepmind}

\pdftrailerid{redacted}

\makeatletter
\renewcommand\bibentry[1]{\nocite{#1}{\frenchspacing\@nameuse{BR@r@#1\@extra@b@citeb}}}
\makeatother

\usepackage{kantlipsum, lipsum}
\usepackage{dsfont}
\usepackage{gdm-colors}
\usepackage{adjustbox}
\usepackage{amsmath}
\usepackage{amssymb}
\usepackage{bm}
\usepackage{colortbl}
\usepackage{comment}
\usepackage{dsfont}
\usepackage{enumitem}
\usepackage{graphicx} 
\usepackage{mathtools}
\usepackage{multicol}
\usepackage{multirow}
\usepackage{lscape}
\usepackage{scalerel}
\usepackage{soul}
\usepackage{subcaption}
\usepackage{tabularx}
\usepackage[most]{tcolorbox}
\usepackage{xspace}
\usepackage{wrapfig}
\usepackage{framed}
\usepackage{parskip}
\usepackage{minted}
\usepackage{appendix} 
\usepackage{cleveref}
\usepackage{algorithm}
\usepackage{algpseudocode}
\usepackage{amsmath}
\usepackage{textcomp}

\usepackage[authoryear, sort&compress, round]{natbib}

\graphicspath{{figures/}}

\title{\textsc{Plantain}: Plan-Answer Interleaved Reasoning}
\correspondingauthor{aliang80@usc.edu}

\reportnumber{001} 

\renewcommand{\today}{}

\author[*,1,2]{Anthony Liang}
\author[1]{Jonathan Berant}
\author[1]{Adam Fisch}
\author[1]{Abhimanyu Goyal}
\author[$\dagger$,1]{Kalpesh Krishna}
\author[$\dagger$,1]{Jacob Eisenstein}

\affil[*]{Lead author}
\affil[$\dagger$]{Co-senior author}
\affil[1]{Google DeepMind}
\affil[2]{University of Southern California}

\newcommand{\namedref}[2]{\hyperref[#2]{#1~\ref*{#2}}}

\newcommand{\smallurl}[1]{\begin{small}\url{#1}\end{small}}

\newcommand{\method}{\textsc{Plantain}\xspace}

\begin{abstract}
Reasoning models often spend a significant amount of time thinking before they generate a visible response. In the meantime, they do not give the user any  hints as to whether their reasoning is on the right track, and do not give the user any recourse  to stop and correct them if their reasoning is  flawed. This creates a frustrating, but unfortunately common, experience: the user's time is wasted while the model reasons from a false premise that could have easily been corrected. In contrast, human speakers typically perform lightweight, incremental grounding acts to ensure that participants in the conversation are on the same page; here we ask if language models can learn to leverage a similar type of behavior? With this motivation, we propose \textit{interleaved reasoning} (IR), in which the model alternates between thinking and surfacing intermediate responses, as an alternative to the standard "think-then-answer" approach. By providing useful information to the user earlier, IR reduces perceived latency, the time a user waits for an initial output, without compromising the quality of the final response.
We further introduce a specialization of interleaved reasoning, \method (\textul{Plan}-\textul{T}hought-\textul{A}nswer \textul{In}terleaving), 
where the first intermediate response is an explicit, step-by-step \textit{plan} for executing the task. This plan-first strategy allows for user intervention and early feedback for subsequent reasoning steps. We demonstrate that \method\ yields an $\sim$6\% improvement in pass@1 across several challenging math reasoning and coding benchmarks, while reducing time-to-first-response by over 60\% relative to think-then-answer baselines.    
\end{abstract}

\begin{document}

\maketitle

\section{Introduction}
\label{sec:introduction}

Reasoning models \citep{jaech2024openai, guo2025deepseek, yang2025qwen3} typically follow a "think-then-answer" paradigm, in which a monolithic, and often quite long, block of reasoning is generated before the model produces any user-facing output. These types of delayed responses not only create a poor user experience, but can often result in substantial time waste when the eventual model response is incorrect. In particular, the "think-then-answer" paradigm offers no opportunity for user intervention: if the model misunderstands an ambiguous prompt or begins its reasoning from a flawed assumption, the user is forced to wait, unaware, while the model pursues an incorrect solution path. This style of black-box reasoning is especially problematic in time-sensitive applications like voice assistants and conversational AI where delayed and irrelevant responses severely impact usability.

In human conversations, this problem is avoided by a range of strategies for \emph{collaborative grounding}, which enable speakers to ensure mutual understanding throughout dialogue~\citep{clark1989contributing,benotti2021grounding,shaikh2023grounding}. Motivated by this literature, we introduce \textit{interleaved reasoning}, in which the model alternates between unobserved "thinking" and surfacing intermediate responses to the user.
We define an intermediate response as a self-contained, usable piece of information that hints at the model's understanding of the user's intention and its plan for satisfying it. 
When this understanding or plan is incorrect, it should be possible for the user to act immediately without waiting for a future revision.
For example, in response to a trip planning prompt, the model might first provide a high-level outline of a possible itinerary. 
Subsequent responses could then provide more granular, day-to-day activities, after making additional web searches and tool calls. 

\begin{figure}[t]
    \centering
    \includegraphics[width=\textwidth]{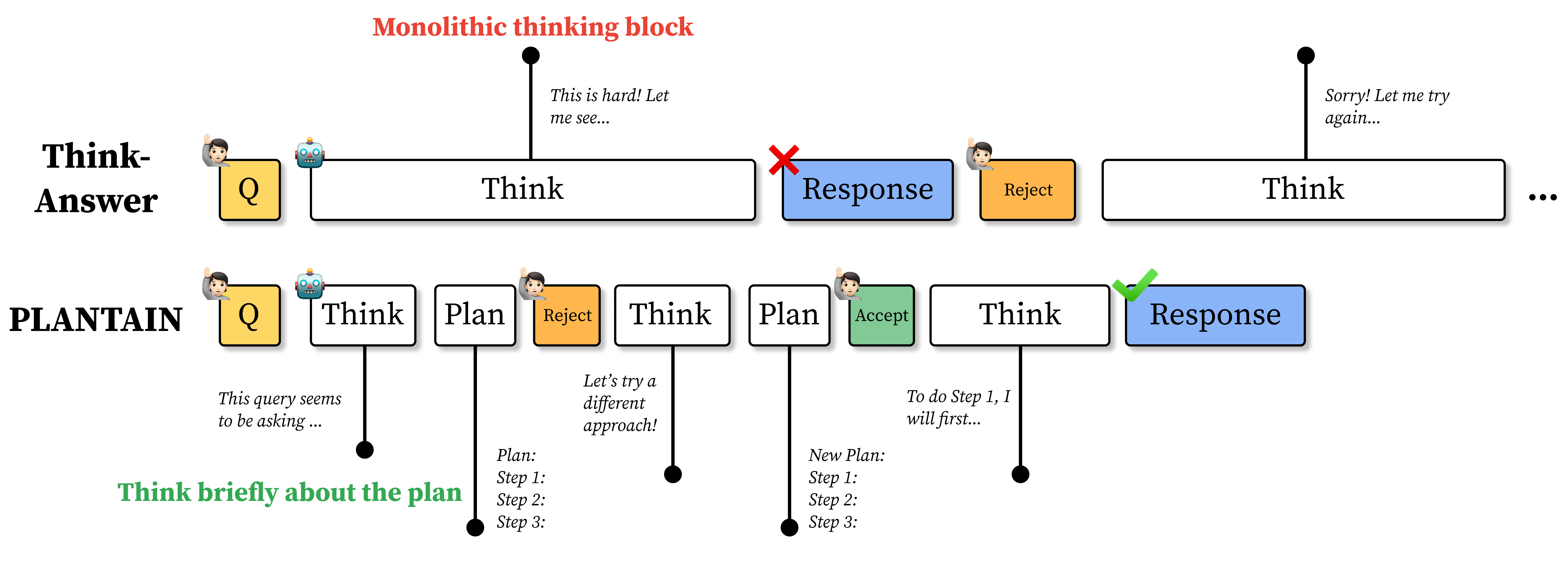}
    \caption{\textbf{\textsc{PLANTAIN} Rewind-and-Repeat} 
After post-training an interleaved reasoning model, we apply an iterative rejection-sampling strategy for inference-time plan guidance. 
\textsc{PLANTAIN} first produces a \emph{plan} as its initial intermediate response.
This plan is evaluated by a human judge or an LLM autorater. 
If accepted, the model proceeds to generate the final answer. 
If rejected, the process is \emph{rewound} such that the rejected plan is appended to a history of failures, and the model is re-prompted to produce a new, distinct plan. 
Crucially, the response generated by \textsc{PLANTAIN} is typically of higher quality than the response produced by a standard Think–Answer model. 
Additionally, subsequent thinking blocks after a rejection plan are short because the model only needs to adjust the plan rather than regenerate full reasoning. 
This leads to early pruning of suboptimal reasoning paths and closer alignment with true user intent.}
    \label{fig:method}
\end{figure}

We propose \method (\textbf{P}lan-\textbf{A}nswer \textbf{I}nterleaved \textbf{R}easoning), a post-training framework that enables existing models to perform interleaved reasoning. 
\method\ follows a three-stage recipe: (1) generating synthetic dataset of interleaved reasoning traces using prompting (2) supervised fine-tuning to distill the desired format into the base model, and (3) reinforcement learning (RL) post-training with verifiable rewards for improving downstream task performance. 
In this paper, we focus on a "plan-first" specialization of interleaved reasoning, shown in \Cref{fig:method}, where the model's initial intermediate response is an explicit, \textit{step-by-step} plan that verbalizes its intended solution path.

While prior work has shown that grounding model generation in a plan improves performance \citep{yao2023react}, our interleaved approach uniquely allows for early intervention to keep the solution on track. 
To this end, we design two inference-time strategies, \textbf{Best-of-N} and \textbf{Rewind-and-Repeat}, that use an LLM-as-a-judge to simulate user feedback on the initial plan(s).
In the Best-of-N approach, the model generates multiple distinct plans, and the user evaluates them simultaneously to select the best plan for the subsequent generation.
In Rewind-and-Repeat, the model proposes one plan at a time, which user can accept or reject; if a plan is rejected, the process is "rewound", and the model is prompted to generate a new, different plan.
We evaluate our method on a diverse set of benchmarks, including math reasoning, coding, text-to-SQL structured generation, and long-context question answering. 
Crucially, our proposed strategies improve alignment and robustness without increasing total token cost, since only short plan prefixes are resampled rather than full generations. 
This allows the model (or user) to prune flawed reasoning paths early, yielding higher pass@1 and lower latency compared to standard Best-of-N sampling over complete responses.

Our core contributions are as follows:
\begin{enumerate}
    \item We introduce \method, a post-training framework for "plan-first" interleaved reasoning, along with two inference-time strategies that allow for early user intervention and feedback.
    \item We demonstrate that our method reduces the time-to-first-token by \textbf{60\%}, which we use as a measure of perceived user latency, without compromising downstream task performance.
    \item We demonstrate that trained exclusively on coding data, \method\ generalizes to diverse reasoning tasks, achieving an average \textbf{+6\%} improvement in pass@1 across all benchmarks.
\end{enumerate}

\section{Related Work}
\label{sec:related_work}

\textbf{LLM Overthinking and Adaptive Thinking.} Reasoning models such as OpenAI O1 \citep{jaech2024openai} and DeepSeekR1 \citep{guo2025deepseek} are trained using RL-based methods to improve task accuracy on complex logic and reasoning tasks. 
Such LLMs notoriously suffer from an overthinking phenomenon where they produce verbose and redundant reasoning steps \citep{chen2024not, sui2025stop}, which not only wastes time, but also can prevents them from providing answers within a given token budget. 
In the extreme, excessive reasoning steps can even eventually introduce severe logical fallacies that lead to incorrect final answers \citep{sui2025stop}.

Several strategies have been proposed to encourage more efficient reasoning. 
One such approach involves integrating a thought length-based reward into the RL framework, penalizing verbose and incorrect answers while encouraging the model to produce more concise reasoning steps \citep{arora2025training, yeo2025demystifying, luo2025o1}.
While effective, these length rewards are typically quite complex, hand-designed and are susceptible to reward hacking.
Another strategy focuses on data curation, where variable length Chain-of-Thought (CoT) reasoning datasets are constructed for Supervised Fine-tuning (SFT) \citep{kang2025c3ot}. 
This is achieved by using heuristics or another LLM to summarize and compress longer reasoning chains without losing key information.
In contrast, we show that training models for interleaved reasoning, alternating between internal thinking and user-facing intermediate responses, naturally encourages concise reasoning without explicit length penalties.
Our approach can effectively parallelize response delivery and reasoning, surfacing useful intermediate outputs while internal thought continues, thereby reducing perceived latency without truncating the reasoning depth or compromising final accuracy.
Finally, \citet{munkhbat2025self} propose to use best-of-N sampling to generate more concise reasoning paths for data curation. 
Our work proposes a related strategy, where instead of using sampling to find the single shortest reasoning path for data curation, we generate multiple plans at inference time, and allow users (more specifically a user simulator, in the context of the experiments in this paper) to select or refine the reasoning trajectory before execution.\looseness=-1

\textbf{Post-hoc Reranking and Inference-Time Steering.} 
Several recent works aim to improve generation quality through inference-time selection or aggregation rather than by modifying the model’s underlying reasoning process. 
Early approaches, such as contrastive decoding \citep{li2022contrastive} and energy-based reranking models \citep{bakhtin2021residual}, follow a similar paradigm, generating multiple candidate continuations and selecting those that best satisfy predefined constraints. 
RANKGEN \citep{krishna2022rankgen} introduces a large encoder trained with contrastive objectives to assess prefix–continuation compatibility, improving coherence and topical relevance when applied to candidate selection. 
RankRAG \citep{yu2024rankrag} unifies reranking and answer generation in retrieval-augmented settings. 
More recently, AGGLM \citep{zhao2025majority} merges reasoning outputs from several trajectories into a unified answer through inference-time aggregation.
While these methods can improve performance, they depend on producing and filtering many reasoning traces after decoding, which is costly.\looseness=-1 

In contrast, our method intervenes directly within the decoding process itself.
Rather than generating full continuations for later rescoring or aggregation, the model produces and evaluates \textit{partial} generations, short plan prefixes, before continuing with the remainder of the response.
Conceptually, this means the unit of generation for evaluation is smaller: instead of scoring entire completions post-hoc, as in RANKGEN or AGGLM, our approach incorporates plan assessment mid-generation.
Unlike speculative decoding, however, this does not require interrupting the decoding process or maintaining multiple concurrent hypotheses.
The interleaved model naturally alternates between planning, reasoning, and answering within a single forward pass.
This shifts the intervention from post-hoc reranking to real-time guidance, enabling efficiency gains without auxiliary reranking stages. 

 


\textbf{Plan Guided Reasoning.} The ReAct framework \citep{yao2023react} interleaves chain-of-thought reasoning with actions allowing the model to incorporate external information sources through tool calls,  
thereby grounding the model and reducing hallucinations in the final answer. 
A follow up work proposes Pre-Act \citep{rawat2025pre}, which creates multi-step execution plan and reasoning for each action. 
In contrast to ReAct-style frameworks, \method produces intermediate answers that are surfaced to the user allowing them to intervene and correct the model's reasoning paths.
Similar to ReAct, we observe that grounding the reasoning in an explicit plan leads to improved downstream accuracy. 
Other prior works have shown that plan-based models, sometimes referred to as \textit{blueprints}, help with model robustness and attribution \citep{gurung2025learning, fierro2024learning}.



\textbf{Interactive LLM Interfaces and Decoding Control.} User-facing LLM interfaces, such as OpenAI’s Deep Research, Gemini’s multi-step reasoning mode, and most standard chat UIs allow users to stop generation mid-thought, truncate rambling explanations, or redirect the model before a full chain of thought is completed.
While effective for improving perceived latency and user control, UI-level interruption does not influence the model’s internal reasoning policy. 
As a result, the model continues to plan to overthink, even if the user frequently stops it. 
From a learning perspective, the system never receives a training signal that shorter or more structured reasoning is preferred. 
In contrast, interleaved reasoning modifies the generation process itself, prompting the model to surface compact, high-value intermediate outputs by default while continuing internal thought as needed. 
\section{Interleaved  Reasoning}
\label{sec:interleaved_reasoning}

We introduce \method, a post-training framework to elicit interleaved reasoning behavior in reasoning models. Unlike prior approaches that impose length restrictions or introduce token budgets \citep{shen2025dast, aggarwal2025l1, han2024token}, \method encourages the model to produce a explicit \textit{plan} as the first intermediate response.
This behavior is controlled through the system instruction (SI), which is modified to prompt the model to “plan first,” while the SFT and RL objectives reinforce the resulting interleaved reasoning style without explicitly constraining when to plan.
Our framework composes a simple three-step recipe: (1) generate synthetic interleaved reasoning trace, (2) supervised fine-tuning to distill the interleaved format, and (3) RL post-training using verifiable rewards. 
After training a model capable of performing interleaved reasoning, we also introduce two inference-time strategies that leverage this output structure to improve the final response generation. 
The complete pseudocode for our training and inference procedures is provided in Appendix~\ref{sec:app:alg}.

\begin{algorithm}[h]
\caption{\method: Post-training to Elicit Interleaved Reasoning}
\label{alg:train-main}
\begin{algorithmic}[1]
\Require Base model $\pi_{\theta_0}$; prompts $\mathcal{X}$; plan-first template $\texttt{SI}_{\text{plan-first}}$; synthetic generator $\Pi^\star$; weights $\alpha_{\text{fmt}}, \alpha_{\text{succ}}$; number of RL updates $K$
\Ensure Interleaved model $\pi_{\theta}$

\State $\mathcal{D}_{\text{interleave}} \!\gets\! \{(x', \tau) \mid x'=\texttt{SI}_{\text{plan-first}}(x),~ \tau \!\sim\! \Pi^\star(\cdot \mid x'),~ x\!\in\!\mathcal{X}\}$ \Comment{Construct synthetic interleaved trace dataset}
\State \textsc{SupervisedFineTune}$(\pi_{\theta_0} \!\rightarrow\! \pi_{\theta};~ \mathcal{D}_{\text{interleave}})$ \Comment{SFT on interleaved traces}

\For{$k = 1$ to $K$} \Comment{RL post-training}
  \State Sample $(x',\cdot)\!\sim\!\mathcal{D}_{\text{interleave}}$; rollout $y\!\sim\!\pi_{\theta}(\cdot\mid x')$
  \State $r\!\gets\!\alpha_{\mathrm{fmt}}\textsc{FormatOK}(y)+\mathbf{1}\{\textsc{FormatOK}(y)=1\}\alpha_{\mathrm{succ}}\textsc{TaskSuccess}(x',y)$
  \State \textsc{PerformRLUpdate}$(\theta; x', y, r)$
\EndFor
\State \Return $\pi_{\theta}$
\end{algorithmic}
\end{algorithm}

\textbf{Interleaved CoT.} A standard language model is typically trained to produce a final answer $A$ conditioned on a prompt $P$, often preceded by a monolithic block of reasoning, $T$, resulting in a sequence of the form, $P \rightarrow T \rightarrow A$. 
Rather than outputting the final answer after a lengthy thought block, interleaved CoT alternates between shorter thoughts and outputting intermediate response to the user.
Formally, an interleaved trace $\tau = (t_1, a_1, t_2, a_2, \dots, t_n, a_n)$ where $t_i$ represents internal thoughts and $a_i$ is an intermediate response surfaced to the user. 

\textbf{Generating Synthetic Interleaved Traces.} Off-the-shelf reasoning models exhibit a strong bias towards the ``think-answer'' paradigm, an artifact of their pretraining that results in lengthy reasoning sequences even for simple prompts \citep{chen2024not}.
Similar to prior works \citep{kang2025c3ot}, we iteratively prompt a larger model to produce variable length CoT traces. 
We take a subset of prompts from \textsc{BigCodeBench} \citep{zhuo2024bigcodebench} and create a natural response decomposition, by modifying the prompt (\Cref{sec:app:dataset_examples}) to ask for a solution outline and unit tests in addition to the code implementation.
Concretely, each interleaved CoT trace follows the structure
\textit{thought $\rightarrow$ solution plan $\rightarrow$ thought $\rightarrow$ code $\rightarrow$ thought $\rightarrow$ unit tests},
encouraging the model to emit useful early outputs for the user.
We provide an example of the generated interleaved traces in Appendix~\ref{sec:app:dataset_examples}.
To improve generalization, we further construct (i) \emph{concatenated-prompt traces}, where multiple independent prompts are concatenated into a composite input, and the model is asked to solve each sequentially within a single reasoning trace, and (ii) \emph{multi-solution traces}, where the model is prompted to generate several distinct candidate solutions for the same problem.
For example, a concatenated-prompt trace may include three standalone coding tasks, such as reversing a string, counting unique elements, and checking for palindromes, encouraging the model to transition between problems within one reasoning context.  
A multi-solution trace could prompt the model to produce multiple implementations for a single task (e.g., recursive and iterative factorial).  
Together, these examples form our synthetic interleaved reasoning dataset, denoted as
$\mathcal{D}_{\text{interleave}} = \{(P_j, \tau_j)\}_{j=1}^{N}$.
This construction diversifies the amount of multi-step interleaving in our training data and prevents the model from overfitting to a fixed output template.


 We fine-tune the base model on $\mathcal{D_{\text{interleave}}}$ by minimizing the negative log-likelihood loss over this dataset. 
This process effectively ``distills'' the desired interactive behavior from our synthetic data into the model, shifting its default response style from monolithic to interleaved.
Afterwards, we post-train the supervised fine-tuned model using Proximal Policy Optimization (PPO) \citep{schulman2017proximal}. PPO utilizes a value network to approximiate the state-value function and Generalized Advantage Estimation to compute the advantage function. 
Prior work \citep{xie2025interleaved} found PPO to be more stable during training compared to GRPO \citep{guo2025deepseek} because of the extra critic model that it requires. The policy model $\pi_{\theta}$ generates rollouts that maximize an expected reward $\mathbb{E}[r(x,y)]$.

\textbf{Reward function.} \label{sec:reward_function} 
For our modified coding prompts, we define a composite reward comprising four rule-based components: 
a \textbf{format} reward that checks whether the response correctly interleaves multiple intermediate answers, 
an \textbf{accuracy} reward based on the pass rate of the generated code against golden unit tests, 
a \textbf{helpfulness} reward produced by an LLM-as-a-judge autorater evaluating the quality of the outline, 
and a \textbf{unit-test} reward indicating whether a valid unit-test block was produced. 
Formally, the overall reward is computed as:
\begin{equation}
    r_{\text{interleave}}(x,y) 
    = r_{\text{format}}(y) \times 
      \big[1 + r_{\text{correctness}}(y) 
      + r_{\text{helpfulness}}(y) 
      + r_{\text{unit\_test}}(y)\big],
\end{equation}
where 
\[
\begin{aligned}
r_{\text{format}}(y) &= \mathbf{1}\{\text{response contains all required sections in the correct order}\},\\
r_{\text{correctness}}(y) &= \frac{\text{\# tests passed}}{\text{\# total tests}},\\
r_{\text{helpfulness}}(y) &= \text{LLM-Judge}(x, y) \in [0, 1],\\
r_{\text{unit\_test}}(y) &= \mathbf{1}\{\text{unit-test block detected}\}.
\end{aligned}
\]
By including the indicator $r_{\text{format}}(y)$ as a multiplicative term, we ensure that downstream rewards are applied only when the interleaved format is satisfied. In preliminary experiments, we find that naively modifying the System Instruction (SI) does not induce the desired interleaved behavior. Models prompted in this way often collapse the plan generation into the reasoning trace and revert to the standard monolithic \textit{think-then-answer} format. This structure is essential for eliciting consistent plan-first reasoning and enabling our subsequent inference-time strategies.

\section{Inference-Time Scaling with Interleaved Plan Generation}
\label{sec:inference_scaling}

Given a trained interleaved reasoning model, we explore inference-time strategies that better align model outputs with user intent and prune erroneous reasoning paths early in the generation process. 
These strategies operate at inference without modifying the underlying model weights, leveraging the model’s ability to produce an explicit \textit{plan} as the first intermediate response.

While we envision these approaches having utility in interactive settings with human users, in this paper we simulate the human using an autorater that provides feedback on candidate plans (full prompt shown in Appendix~\ref{app:sec:autorater}). 
The judge receives as input the user’s original prompt $p$ and a proposed plan $a_i$; because it has access to the user prompt and plan but not the model’s internal reasoning trace, it evaluates plans solely from the perspective of an external user.
Given $N$ candidate plans $\{p_i\}_{i=1}^N$, the judge is prompted to either (i) select the index of the best plan when $N>1$, or (ii) output a binary decision $\{\texttt{accept}, \texttt{reject}\}$ when evaluating a single plan ($N=1$).  

Given that our model can interleave between internal thinking and user-facing responses, we use the first intermediate output as a \textit{plan}.  
We introduce two inference-time control methods based on this plan structure:  
(1) \emph{Best-of-$N$}, which samples multiple candidate plans and uses the LLM autorater to select the plan that best addresses the user prompt, and  
(2) \emph{Iterative Plan Rejection Sampling}, which generates a new plan whenever the current one is rejected, continuing until an acceptable plan is found or a retry budget is exhausted.  
Together, these strategies enable adaptive, user-aligned reasoning at inference time without retraining.  
Although the model is initially trained on coding tasks, these inference-time mechanisms are domain-agnostic and extend naturally to other reasoning settings such as mathematics, text understanding, or planning.
Pseudocode for the two inference-time strategies are provided in Algorithm~\ref{alg:bestof_main} and \ref{alg:reject_main}.

\begin{algorithm}[h]
\caption{Inference Strategy 1: Best-of-$N$ Plan}
\label{alg:bestof_main}
\begin{algorithmic}[1]
\Require Interleaved model $\pi_{\theta}$; prompt $p$; temperature $\tau$; number of plans $N$; AutoRater $R$
\State $p' \!\gets\! \texttt{SI}_{\text{plan-first}}(p)$ \Comment{convert to plan-first prompt}
\State $\{(a_i, s_i)\}_{i=1}^{N} \!\gets\! \{(a_i, R(p, a_i)) \mid a_i \!\sim\! \pi_{\theta}(a \mid p'; \tau)\}$ \Comment{sample and score $N$ plans}
\State $a^\star \!\gets\! \arg\max_{a_i} s_i$ \Comment{select best plan}
\State $y \!\sim\! \pi_{\theta}(\cdot \mid p', a^\star)$ \Comment{roll out reasoning and answer}
\State \Return $y$
\end{algorithmic}
\end{algorithm}

\begin{algorithm}[h]
\caption{Inference Strategy 2: Rewind \& Repeat}
\label{alg:reject_main}
\begin{algorithmic}[1]
\Require Interleaved model $\pi_{\theta}$; prompt $p$; retries $T$; AutoRater $R$
\State $p' \!\gets\! \texttt{SI}_{\text{plan-first}}(p)$;\quad $\mathcal{H} \!\gets\! \varnothing$ \Comment{initialize prompt and history}
\For{$t = 1$ to $T$}
  \State $a_t \!\sim\! \pi_{\theta}(a \mid \text{Augment}(p', \mathcal{H}))$;\; $d_t \!\gets\! R(p, a_t)$ \Comment{propose and rate plan}
  \If{$d_t = \texttt{accept}$}
    \State \Return $y \!\sim\! \pi_{\theta}(\cdot \mid p', a_t)$ \Comment{accept and complete reasoning}
  \Else
    \State $\mathcal{H} \!\gets\! \mathcal{H} \cup \{a_t\}$ \Comment{store rejected plan}
  \EndIf
\EndFor
\State $a_r \!\sim\! \text{Uniform}(\mathcal{H})$; \quad \Return $y \!\sim\! \pi_{\theta}(\cdot \mid p', a_r)$ \Comment{fallback rollout}
\end{algorithmic}
\end{algorithm}

\textbf{Best-of-N Selection.}
In this setting, the model generates a diverse set of $N$ candidate plans, $\{a_1, a_2, \dots, a_N\}$, for a given user prompt $p$.
To encourage diversity, we sample with a high temperature parameter ($\tau > 1$), which increases the likelihood of exploring less frequent but potentially insightful reasoning paths. The generated plans are then evaluated by the LLM \textit{autorater}, which serves as a proxy for human feedback.
The autorater receives the user’s prompt $p$ along with the set of candidate plans, but not the model’s internal reasoning process, introducing \textit{information asymmetry} between the model and autorater.  
This setup mirrors real user evaluation, where one can judge whether a plan is well-structured and relevant to the prompt without access to the model’s internal thoughts.  
The autorater selects the plan that best addresses the prompt and is most likely to lead to a correct or complete final response, denoted as $a^\star$.  
The model then continues generation conditioned on $a^\star$, producing the subsequent reasoning steps and final answer.

\textbf{Iterative Plan Generation with Rejection Sampling.}
Instead of generating multiple plans at once, this approach performs plan generation in an iterative feedback loop.  
The model first proposes an initial plan $a_1$ conditioned on the user prompt $p$.  
The LLM autorater then evaluates the plan and returns a binary verdict $d_1 \in \{\texttt{accept}, \texttt{reject}\}$.  
If accepted, the model proceeds to complete the response using $a_1$.  
If rejected, the plan is added to the failure set $\mathcal{H} = \mathcal{H} \cup \{a_1\}$, and the model is re-prompted with the original query and rejection history.  
This conditioning encourages the model to generate a new plan that avoids previous failure modes.  
The process repeats for up to $T$ iterations or until a plan is accepted.  
If all attempts are rejected, the model samples a fallback plan from $\mathcal{H}$ and continues generation from that plan.

\section{Experiment Setup}
\label{sec:experiments}

We aim to study the efficacy of each component of \method and compare it against baselines. 
To this end, we organize our experiments to answer the following questions:

\begin{enumerate}[label=(\textbf{Q\arabic*}), nosep]
    \item Does interleaved reasoning reduce inference-time \textit{latency} without compromising the final task performance compared to the standard think-answer approach? 
    \item How effective are our inference-time strategies at improving initial plan quality and pruning suboptimal reasoning paths before execution?
\end{enumerate}

\textbf{Models and Baselines.} We use the \textsc{Qwen3} model family \citep{yang2025qwen3}, an open-source reasoning model with built-in thinking capabilities, as our base architecture, evaluating both 4B and 8B parameter variants.  
We compare our post-trained interleaved reasoning models and inference-time strategies against several baselines and ablations.  
The first set of baselines are applied directly to the base model without any additional fine-tuning.
\begin{itemize}
    \item \textbf{No Thinking:} The base model is prompted to generate the answer directly, without explicit chain-of-thought reasoning.
    \item \textbf{Think-Answer (TA):} The base model is prompted to perform explicit reasoning steps in a monolithic block before generating the final answer.
    \item \textbf{Rewind-and-Repeat (R\&R) on Answer:} Rewind-and-repeat applied to the \textit{final answer}. 
    An LLM autorater judges the final response and then trigger a full restart from the original prompt if the answer is rejected. This inference strategy is equivalent to our proposed $R\&R$, but applied at the final response level.
\end{itemize}

We compare the baselines to our inference-time strategies applied on the RL-trained interleaved reasoning model.
\begin{itemize}
    \item \textbf{Plan-Answer:} Direct inference of the interleaved thinking model which first generates a plan and then the final answer.
    \item \textbf{Best-of-N Plan:} RL fine-tuned model, generate $N$ diverse plans in parallel, using an LLM autorater to select the best plan upon which final response is conditioned 
    \item \textbf{Rewind-and-Repeat (R\&R) Plan:} Iterative rejection-sampling approach that generates single plan at a time, with an LLM autorater to decide whether to accept it and continue, or reject and "rewind" to generate a new plan 
\end{itemize}

\textbf{Interleaved and Evaluation Datasets.}
We train on a combination of coding and mathematical reasoning datasets (BigCodeBench, MBPP, and MATH500) and evaluate across broader domains including text-to-SQL and long-context question answering to assess cross-domain generalization.
BigCodeBench~\citep{zhuo2024bigcodebench} contains 1.1K function-level Python coding tasks requiring multi-library reasoning and compositional code synthesis.
MBPP~\citep{austin2021program} consists of 974 entry-level Python problems designed for evaluating basic programming competence.
MATH500~\citep{lightman2023lets} includes 500 symbolic reasoning problems spanning algebra, geometry, and probability.
BirdSQL~\citep{li2023can} and QuALITY~\citep{pang2021quality} are used for out-of-domain evaluation: the former tests text-to-SQL translation grounded in real relational schemas, while the latter measures long-context reading comprehension with passages averaging 5K tokens.
See Appendix~\ref{sec:app:dataset_deats} for further dataset and split details.
We generate interleaved traces on 50 BigCodeBench prompts modified to request a solution outline and unit tests in addition to the code solution.
Iterative CoT prompting on \textsc{Qwen3-32B} is used to synthesize these traces.
We additionally collect 50 traces by concatenating pairs of MBPP prompts and 25 traces that request multiple solutions for a single coding problem, resulting in 125 total interleaved reasoning traces used for SFT.
 
\textbf{Evaluation Metrics.}  
We evaluate model performance using four primary metrics: \textbf{task success rate (pass@1)}, \textbf{unit test pass rate}, \textbf{time-to-first-response (TTFR)}, and \textbf{tokens-per-problem (T/P)}.  
Pass@1 measures the percentage of prompts for which the model produces a fully correct solution on the first attempt and is reported across all benchmarks.  
For coding tasks, we additionally report the unit test pass rate, the proportion of ground-truth unit tests passed by the generated code, normalized by the total number of tests.  
TTFR captures the number of \textit{thought tokens} generated before the first user-visible response (i.e., the initial plan), providing a measure of response latency.
While token count does not map linearly to wall-clock time, it provides a reliable model-agnostic proxy for response latency, since inference speed is approximately proportional to the number of generated tokens under a fixed decoding setup.
The T/P ratio quantifies overall token efficiency as the average number of tokens generated per problem, including both internal reasoning and user-facing outputs.  
For interleaved traces, we further assess intermediate response quality using an LLM-as-a-judge (\textsc{Qwen2.5-7B-Instruct}), which labels each segment as \textit{helpful} or \textit{not helpful}.

For our inference-time strategies, we compute TTFR based on the first generated plan, even if it is later rejected by the autorater. 
When reporting T/P, we include all thought tokens from any additional plan generations. 
In the \textit{Best-of-$N$} setting, plans are generated in parallel, so latency is bounded by the slowest plan. 
In contrast, the \textit{Rewind \& Repeat (R\&R)} strategy generates and evaluates plans sequentially, though the short length of each thought block results in negligible overhead relative to the full rollout.

\section{Results}

\noindent\textbf{Interleaved reasoning generalizes beyond coding, reducing latency and improving task accuracy across domains.}  
Table~\ref{tab:ir} compares Think-Answer (TA) and Plan-Answer decoding for \textsc{Qwen3-4B} and \textsc{Qwen3-8B} across MATH500, MBPP, Text-to-SQL, and QuaLITY.  
Models trained to interleave planning with reasoning on coding data transfer effectively to unseen domains, substantially lowering time-to-first-response (TTFR) while maintaining or improving final task performance.  
For example, on MATH500, TTFR decreases from $2044 \rightarrow 628$ tokens for \textsc{Qwen3-4B} ($84.2 \rightarrow 84.4$ P@1) and from $2106 \rightarrow 625$ tokens for \textsc{Qwen3-8B} ($88.2 \rightarrow 85.2$ P@1), with similar trends on MBPP and QuaLITY where TTFR is reduced by over 60\%.  
The ``No Thinking'' baseline yields TTFR = 0 because the model directly outputs an answer without generating any intermediate reasoning tokens; this serves as a lower bound on perceived latency but typically produces less reliable outputs.  
Despite being post-trained only on coding tasks, the model exhibits consistent improvements on math, text-to-SQL, and reading comprehension, indicating that explicit planning structures learned in one domain promote more efficient and accurate reasoning in others.  
This demonstrates that interleaved reasoning not only reduces pre-answer token overhead but also enhances generalization and response quality across diverse reasoning tasks.

\begin{table*}[h!]
\centering
\small
\setlength{\tabcolsep}{4pt}
\begin{tabular}{@{}l|cc|cc|cc|cc|cc@{}}
\toprule
\multirow{2}{*}{\textbf{Method}} &
\multicolumn{2}{c|}{\textbf{MATH500}} &
\multicolumn{2}{c|}{\textbf{MBPP}} &
\multicolumn{2}{c|}{\textbf{Text-to-SQL}} &
\multicolumn{2}{c|}{\textbf{QuaLITY}} &
\multicolumn{2}{c}{\textbf{Average}} \\
\cmidrule(lr){2-11}
& \textbf{P@1$\uparrow$} & \textbf{TTFR$\downarrow$} &
\textbf{P@1$\uparrow$} & \textbf{TTFR$\downarrow$} &
\textbf{P@1$\uparrow$} & \textbf{TTFR$\downarrow$} &
\textbf{P@1$\uparrow$} & \textbf{TTFR$\downarrow$} &
\textbf{P@1$\uparrow$} & \textbf{TTFR$\downarrow$} \\
\midrule

\rowcolor{gray!10}
\multicolumn{11}{l}{\textsc{Qwen3-4B}} \\[1pt]
No Thinking & 81.6 & 0 & 48.2 & 0 & 18.6 & 0 & 61.0 & 0 & 52.4 & 0 \\
Think-Answer & 82.8 & 1492 & 50.8 & 1298 & 25.4 & 1542 & 68.4 & 1345 & 56.9 & 1419 \\
\textbf{Plan-Answer} & \textbf{84.4} & \textbf{628} & \textbf{52.6} & \textbf{523} & \textbf{26.0} & \textbf{484} & \textbf{72.2} & \textbf{428} & \textbf{58.8} & \textbf{516} \\

\midrule
\rowcolor{gray!10}
\multicolumn{11}{l}{\textsc{Qwen3-8B}} \\[1pt]
No Thinking & 82.4 & 0 & 51.4 & 0 & 21.0 & 0 & 64.2 & 0 & 54.8 & 0 \\
Think-Answer & 83.0 & 1587 & 54.5 & 1417 & 28.0 & 1684 & 70.6 & 1463 & 59.0 & 1538 \\
\textbf{Plan-Answer} & \textbf{85.2} & \textbf{625} & \textbf{56.8} & \textbf{554} & \textbf{28.0} & \textbf{457} & \textbf{76.4} & \textbf{443} & \textbf{61.6} & \textbf{520} \\

\bottomrule
\end{tabular}
\caption{\textbf{Interleaved reasoning generalizes beyond coding tasks, reducing latency and preserving accuracy.} 
Plan-first decoding (\textit{Plan-Answer}) substantially lowers time-to-first-response (TTFR) while maintaining or improving pass@1 accuracy compared to the standard \textit{Think-Answer} baseline with a token budget of 4096. 
Despite being post-trained only on coding data, the learned interleaved behavior transfers effectively to math, text-to-SQL, and reading comprehension tasks.}
\label{tab:ir}
\end{table*}

\vspace{1em}
\noindent\textbf{Early plan-level feedback guides reasoning toward correct solution paths, yielding up to +2–3\% higher task accuracy with $7\times$ lower TTFR.}
Table \ref{tab:inference_scaling} compares inference-time control strategies on the RL-trained interleaved model.
Unlike answer-level feedback that arrives only after full reasoning, plan-level evaluation intervenes \emph{before} subsequent reasoning begins, allowing the model to revise faulty plans and steer downstream thoughts toward more accurate outcomes.
Both Best-of-$N$ and Rewind-and-Repeat (R\&R) at the plan stage outperform the base Plan-Answer decoding, showing that lightweight, inference-time guidance alone improves reasoning quality without retraining.
For \textsc{Qwen3-4B}, R\&R (Plan) boosts MATH500 accuracy from $84.4 \rightarrow 86.8$ while maintaining $\sim$578 TTFR tokens, over $7\times$ faster than R\&R (Answer).
Similarly, \textsc{Qwen3-8B} achieves $+2.2$ P@1 on MATH500 and $+0.6$ on MBPP with TTFR $\approx600$.
These results demonstrate that plan interleaving provides earlier, denser supervision that improves both efficiency and final task accuracy.

\begin{table*}[h!]
\centering
\small
\setlength{\tabcolsep}{4pt}
\begin{tabular}{@{}l|cc|cc|cc|cc|cc@{}}
\toprule
\multirow{2}{*}{\textbf{Method}} &
\multicolumn{2}{c|}{\textbf{MATH500}} &
\multicolumn{2}{c|}{\textbf{MBPP}} &
\multicolumn{2}{c|}{\textbf{Text-to-SQL}} &
\multicolumn{2}{c|}{\textbf{QuaLITY}} &
\multicolumn{2}{c}{\textbf{Average}} \\
\cmidrule(lr){2-11}
& \textbf{P@1$\uparrow$} & \textbf{TTFR$\downarrow$} &
\textbf{P@1$\uparrow$} & \textbf{TTFR$\downarrow$} &
\textbf{P@1$\uparrow$} & \textbf{TTFR$\downarrow$} &
\textbf{P@1$\uparrow$} & \textbf{TTFR$\downarrow$} &
\textbf{P@1$\uparrow$} & \textbf{TTFR$\downarrow$} \\
\midrule

\rowcolor{gray!10}
\multicolumn{11}{l}{\textsc{Qwen3-4B}} \\[1pt]
Plan-Answer & 84.4 & 628 & 52.6 & 523 & 26.0 & 484 & 72.2 & 428 & 58.8 & 516 \\
R\&R (Answer) & 84.4 & 4079 & 52.4 & 3321 & 31.2 & 2822 & 74.0 & 2453 & 60.5 & 3169 \\
Best-of-N (5) & 86.2 & \textbf{523} & \textbf{54.6} & 584 & 28.4 & 568 & \textbf{75.2} & 547 & 61.1 & 556 \\
\textbf{R\&R (Plan)} & \textbf{86.8} & 578 & 53.8 & \textbf{519} & \textbf{31.6} & \textbf{552} & 75.0 & \textbf{534} & \textbf{61.8} & \textbf{546} \\

\midrule
\rowcolor{gray!10}
\multicolumn{11}{l}{\textsc{Qwen3-8B}} \\[1pt]
Plan-Answer & 85.2 & 625 & 56.8 & 554 & 28.0 & 457 & 76.4 & 443 & 61.6 & 520 \\
R\&R (Answer) & 86.0 & 4279 & 56.2 & 3541 & 34.0 & 2957 & 78.2 & 2614 & 63.6 & 3348 \\
Best-of-N (5) & 85.8 & 628 & \textbf{58.2} & 641 & 31.0 & 602 & 78.2 & 582 & 63.3 & 613 \\
\textbf{R\&R (Plan)} & \textbf{89.4} & \textbf{611} & 57.4 & \textbf{543} & \textbf{34.0} & \textbf{568} & 78.2 & \textbf{562} & \textbf{64.8} & \textbf{571} \\

\bottomrule
\end{tabular}
\caption{\textbf{Inference-time strategies:}
Plan-level feedback provides earlier and more informative supervision than answer-level feedback, yielding higher accuracy and lower TTFR.
Both Best-of-N and R\&R (Plan) improve over the base Plan-Answer decoding without additional training.}
\label{tab:inference_scaling}
\end{table*}

\subsection{Ablations and Analysis}
\label{sec:ablations}

\begin{table}[h!]
\centering
\small
\setlength{\tabcolsep}{4pt}
\begin{tabular}{@{}l|c|c|c|c|c@{}}
\toprule
\textbf{Method (4096 $\rightarrow$ 8192)} &
\textbf{MATH500} &
\textbf{MBPP} &
\textbf{Text-to-SQL} &
\textbf{QuaLITY} &
\textbf{Average} \\
\midrule

\rowcolor{gray!10}
\multicolumn{6}{l}{\textsc{Qwen3-4B}} \\[1pt]
Think-Answer (TA) & 82.8 $\rightarrow$ 84.2 & 50.8 $\rightarrow$ 52.0 & 25.4 $\rightarrow$ 28.2 & 68.4 $\rightarrow$ 74.6 & 56.9 $\rightarrow$ 59.8 \\
Plan-Answer & 84.4 $\rightarrow$ 86.0 & 52.6 $\rightarrow$ 53.5 & 26.0 $\rightarrow$ 29.5 & 72.2 $\rightarrow$ 74.8 & 58.8 $\rightarrow$ 60.9 \\
R\&R (Plan) & 86.8 $\rightarrow$ 87.2 & 53.8 $\rightarrow$ 54.3 & 31.6 $\rightarrow$ 31.7 & 75.0 $\rightarrow$ 75.4 & \textbf{61.8 $\rightarrow$ 62.2} \\

\midrule
\rowcolor{gray!10}
\multicolumn{6}{l}{\textsc{Qwen3-8B}} \\[1pt]
Think-Answer (TA) & 83.0 $\rightarrow$ 88.2 & 54.5 $\rightarrow$ 56.4 & 28.0 $\rightarrow$ 31.0 & 70.6 $\rightarrow$ 78.4 & 59.0 $\rightarrow$ 63.5 \\
Plan-Answer & 85.2 $\rightarrow$ 88.8 & 56.8 $\rightarrow$ 58.5 & 28.0 $\rightarrow$ 32.5 & 76.4 $\rightarrow$ 78.2 & 61.6 $\rightarrow$ 64.5 \\
R\&R (Plan) & 89.4 $\rightarrow$ 89.6 & 57.4 $\rightarrow$ 59.1 & 34.0 $\rightarrow$ 34.3 & 78.2 $\rightarrow$ 78.4 & \textbf{64.8 $\rightarrow$ 65.4} \\

\bottomrule
\end{tabular}
\caption{\textbf{Token-budget ablation:} Interleaved inference achieves near-saturated accuracy at 4K tokens, highlighting efficient use of available context. 
While Think-Answer continues to improve with longer reasoning—reflecting redundant token usage—Plan-Answer benefits modestly from larger budgets but remains below Rewind-and-Repeat, indicating stable yet non-adaptive scaling. 
Rewind-and-Repeat achieves the strongest token efficiency and overall accuracy.}
\label{tab:token_budget}
\end{table}

\noindent\textbf{Think-Answer models rely on longer reasoning chains to improve, indicating inefficient token usage, whereas interleaved models attain comparable accuracy with fewer tokens.}
Table~\ref{tab:token_budget} varies the available context window ($4096 \rightarrow 8192$ tokens).
While Think-Answer (TA) models show only modest gains with larger budgets (e.g., MATH500 $82.8 \rightarrow 84.2$), interleaved models nearly saturate at 4096 tokens ($86.8 \rightarrow 87.2$ for \textsc{Qwen3-4B}; $89.4 \rightarrow 89.6$ for \textsc{Qwen3-8B}).
This demonstrates that 4K tokens are sufficient for plan-conditioned inference, even without extended “thinking” budgets.

In contrast, the TA baseline exhibits poor token efficiency. Its improvements come primarily from longer reasoning sequences, much of which is spent on redundant or self-corrective thought.
Empirically, many TA generations fail to terminate naturally and in these cases, the model continues reasoning until truncated, after which a final answer must be explicitly prompted.
Interleaved models, by comparison, allocate tokens adaptively and terminate after a few reasoning–answer alternations, effectively regularizing total reasoning length.

In the \textit{Rewind-and-Repeat} setting, this difference becomes particularly salient. Because interleaved models surface an explicit plan before full reasoning, each rewind operates over a concise, interpretable intermediate representation rather than an entire free-form completion.
This allows the LLM-as-a-judge to prune low-quality trajectories early—avoiding wasted computation and ensuring that subsequent reasoning unfolds around a verified plan, whereas TA models can only rewind after full generations, offering no opportunity for early correction.
See Appendix~\ref{sec:app:math500_case_study} for qualitative traces illustrating these dynamics.

On MATH500, for example, the TA model frequently degenerates into repetitive or self-contradictory reasoning and fails to recover once diverged, while plan-based interleaving grounds the process through explicit subgoals that constrain and stabilize subsequent thoughts.
Overall, these results show that interleaved inference regularizes reasoning depth, avoids unnecessary computation, and achieves higher token efficiency under fixed or limited context budgets.



\begin{table}[h!]
\centering
\small
\setlength{\tabcolsep}{4pt}
\begin{tabular}{lccccc}
\toprule
\textbf{Metric} & \textbf{MATH500} & \textbf{MBPP} & \textbf{Text-to-SQL} & \textbf{QuaLITY} & \textbf{Average} \\
\midrule
Plans needing rewind & 120/500 & 55/257 & 60/250 & 40/250 & 22.4\% \\
First rewind approved & 60.8\% & 54.5\% & 63.3\% & 67.5\% & 61.5\% \\
Second rewind approved & 25.8\% & 18.2\% & 21.7\% & 17.5\% & 20.8\% \\
Rewounded P@1 & 86.6\% & 72.7\% & 85.0\% & 82.0\% & 81.6\% \\
\bottomrule
\end{tabular}
\caption{\textbf{Rewind statistics.} Fraction of prompts requiring plan rewinds and corresponding approval rates. 
On average, only 22\% of plans require a rewind, and over 80\% of corrections succeed within two attempts, demonstrating the efficiency of plan-level R\&R.}
\label{tab:rewind_stats}
\end{table}

\noindent\textbf{A small number of plan-level interventions recover most failure cases, highlighting the efficiency of structured feedback.}  
Table~\ref{tab:rewind_stats} analyzes how often plan-level intervention is needed and how quickly the model converges under the Rewind \& Repeat (R\&R) strategy. 
Across all benchmarks, only 20–25\% of plans require a rewind, and the majority of those are corrected on the first retry (60–65\%). 
A small fraction (15–25\%) benefit from a second rewind, after which success rates exceed 80–85\%. 
These results indicate that most initial plans are already well-formed and R\&R primarily serves to re-ground reasoning early, preventing the model from pursuing unproductive solution paths.  
This mechanism contributes directly to the token efficiency observed in Table~\ref{tab:inference_scaling}, where plan-level feedback reduces reasoning depth without sacrificing accuracy.  
Qualitatively, we find that rewinds promote more stable and targeted reasoning trajectories, allowing the model to converge faster on valid solutions with minimal additional tokens.

\section{Conclusion}
\label{sec:conclusion}

We introduced \method, a post-training framework that elicits \textit{interleaved reasoning} behavior in large reasoning models.  
By encouraging models to first produce an explicit plan and then alternate between internal reasoning and user-facing responses, \method reduces latency and improves controllability.  
Empirically, interleaved reasoning achieves up to $7\times$ lower time-to-first-response (TTFR) and $+2$–$3\%$ higher task accuracy compared to standard \textit{Think-Answer} decoding, while maintaining comparable total token usage.  
Our interleaved model learns to structure its reasoning process naturally, without explicit length penalties or handcrafted constraints.  
At inference time, we introduced two plan-level control strategies, Best-of-$N$ selection and Rewind \& Repeat, that further improve response quality and efficiency without additional training.  
Although trained solely on code generation tasks, the resulting model generalizes effectively to broader reasoning domains such as math and long-context question answering.  

\method focuses on structured, verifiable tasks where plan correctness can be automatically assessed.  
An important next step is to conduct human studies evaluating the perceived usefulness, interpretability, and responsiveness of interleaved reasoning in interactive settings.  
We also plan to extend evaluations to scenarios with ambiguous or under-specified user prompts, where users can provide early feedback on intermediate plans to guide the model toward their true intent.  
Finally, the LLM-as-a-judge used for plan evaluation introduces nontrivial latency and may reflect biases of the underlying model.
Future work could mitigate these effects by incorporating human preference data or lightweight learned reward models to better calibrate plan evaluation and align judgments with user intent.

\section*{Acknowledgments}

We are deeply grateful to Ting-yun Chang, Frederick Zhang, and Ming Zhong, Alexandra Chronopoulou, Xiang Zhou, and Shyam Upadhyay for the discussions and thoughtful feedback that helped refine the methodology, experimental design, and paper.

\bibliographystyle{abbrvnat}
\nobibliography*
\bibliography{references}

\appendix
\begin{appendices}
\newpage

\section{Appendix}
\subsection{\method Algorithm}
\label{sec:app:alg}

\begin{algorithm}[h]
\caption{\method: Post-training to Elicit Interleaved Reasoning}
\label{alg:train}
\begin{algorithmic}[1]
\Require Base LM $\pi_{\theta_0}$; reference policy $\pi_{\text{ref}}$; value net $V_\phi$;
         prompts $\mathcal{X}$; SI template $\texttt{SI}_\text{plan-first}$;
         synthetic generator $\Pi^\star$ (larger LM);
         weights $\alpha_{\text{fmt}}, \alpha_{\text{acc}}, \alpha_{\text{help}}, \alpha_{\text{ut}}$; PPO coeff $\beta$
\Ensure Interleaved model $\pi_{\theta}$ (plan-first, then alternating thoughts/answers)

\Statex \textbf{Synthetic Interleaved Trace Dataset}
\State $\mathcal{D}_{\text{interleave}} \gets \varnothing$
\ForAll{$x \in \mathcal{X}$}
  \State $x' \gets \texttt{SI}_\text{plan-first}(x)$
  \State $\tau = (t_1,a_1,t_2,a_2,\ldots,t_n,a_n) \sim \Pi^\star(\cdot \mid x')$ \Comment{$a_1$ is the explicit \emph{plan}}
  \State $\mathcal{D}_{\text{interleave}} \gets \mathcal{D}_{\text{interleave}} \cup \{(x', \tau)\}$
\EndFor

\Statex \textbf{Supervised Fine-tuning}
\State $\theta \gets \arg\min_{\theta} \sum_{(x',\tau)\in \mathcal{D}_{\text{interleave}}} \big(-\log \pi_{\theta}(\tau \mid x')\big)$
\Comment{Shift style from monolithic to interleaved}

\Statex \textbf{RL Post-Training (PPO)}
\Repeat
  \State Sample $(x',\cdot) \sim \mathcal{D}_{\text{interleave}}$; rollout $y \sim \pi_{\theta}(\cdot \mid x')$
  \State $r_{\mathrm{fmt}} \gets \textsc{FormatOK}(y)$ \Comment{plan-first; valid interleaving; required sections}
  \State $r_{\mathrm{acc}} \gets \textsc{UnitTestPassRate}(y)$
  \State $r_{\mathrm{help}} \gets \textsc{LLMJudgeHelpfulness}(x',y)$
  \State $r_{\mathrm{ut}} \gets \mathbf{1}\{\text{unit\_tests present in } y\}$
  \State $g \gets \mathbf{1}\{r_{\mathrm{fmt}}=1\}$ \Comment{gate downstream rewards on correct interleaved format}
  \State $r(x',y) \gets \alpha_{\mathrm{fmt}} r_{\mathrm{fmt}} + g\cdot\big(\alpha_{\mathrm{acc}} r_{\mathrm{acc}} + \alpha_{\mathrm{help}} r_{\mathrm{help}} + \alpha_{\mathrm{ut}} r_{\mathrm{ut}}\big)$
  \State Compute advantages $\hat{A}$ with GAE using $V_\phi$; update $\theta,\phi$ via PPO:
  \State \hskip1em $\displaystyle \max_{\theta}\; \mathbb{E}\!\left[\text{clip}\!\left(\frac{\pi_{\theta}}{\pi_{\theta_{\text{old}}}},\,1\!\pm\!\epsilon\right)\hat{A}\right]
                      \;-\; \beta\,\mathrm{D_{KL}}\!\big(\pi_{\theta}(\cdot\mid x')\,\|\,\pi_{\text{ref}}(\cdot\mid x')\big)$
\Until{convergence}
\State \Return $\pi_{\theta}$
\end{algorithmic}
\end{algorithm}

\begin{algorithm}[h]
\caption{Inference Strategy 1: Best-of-$N$ Plan}
\label{alg:bestof}
\begin{algorithmic}[1]
\Require Trained interleaved model $\pi_{\theta}$; prompt $p$; temperature $\tau>1$; number of plans $N$; LLM Autorater $R$
\State Form plan-first instruction $p' \gets \texttt{SI}_\text{plan-first}(p)$
\State Generate candidate plans $\mathcal{A} \gets \{a_i \sim \pi_{\theta}(a \mid p'; \tau)\}_{i=1}^{N}$
\State Score each plan $s_i \gets R(p, a_i)$
\State Select $a^\star \gets \arg\max_{a_i \in \mathcal{A}} s_i$
\State Roll out remainder conditioned on $a^\star$: $y \sim \pi_{\theta}(\cdot \mid p', a^\star)$
\State \Return $y$
\end{algorithmic}
\end{algorithm}

\begin{algorithm}[h]
\caption{Inference Strategy 2: Rewind \& Repeat - Iterative Plan Rejection Sampling}
\label{alg:reject}
\begin{algorithmic}[1]
\Require Trained interleaved model $\pi_{\theta}$; prompt $p$; maximum retries $T$; LLM Autorater $R$
\State $p' \gets \texttt{SI}_\text{plan-first}(p)$;\quad $\mathcal{H} \gets \varnothing$
\For{$t = 1$ to $T$}
  \State $c_t \gets \text{Augment}(p', \mathcal{H})$
  \State Propose plan $a_t \sim \pi_{\theta}(a \mid c_t)$
  \State $d_t \gets R(p, a_t)$
  \If{$d_t = \texttt{accept}$}
     \State \Return $y \sim \pi_{\theta}(\cdot \mid c_t, a_t)$
  \Else
     \State $\mathcal{H} \gets \mathcal{H} \cup \{a_t\}$
  \EndIf
\EndFor
Sample a random plan $a_r \sim \text{Uniform}(\mathcal{H})$
\State \Return $y \sim \pi_{\theta}(\cdot \mid p', a_r)$
\end{algorithmic}
\end{algorithm}

\newpage 

\subsection{Interleaved Reasoning System Instruction}
\label{sec:app:interleave_si}
\noindent\rule{\textwidth}{1pt}

You are a helpful assistant. You reason through problems step-by-step before providing an answer. You conduct your reasoning within \texttt{<think>}\texttt{</think>} and share partial answers that are useful for the user within \texttt{<answer>}\texttt{</answer>}. You continue this pattern of \texttt{<think>}...\texttt{</think>} \texttt{<answer>}...\texttt{</answer>} \texttt{<think>}...\texttt{</think>} \texttt{<answer>}...\texttt{</answer>} until you reach the final answer.

User: \{insert prompt here\} 

Assistant: 

\noindent\rule{\textwidth}{1pt}












\newpage 

\subsection{LLM Plan Autorater}
\label{app:sec:autorater}

\noindent\rule{\textwidth}{1pt}
You are an expert plan evaluator. Given a prompt and a single plan, your task is to determine if the plan is good enough to proceed with.

Evaluation Criteria:

\begin{enumerate}
    \item Completeness: Does the plan address all aspects of the prompt?
    \item Feasibility: Is the plan realistic and implementable?
    \item Clarity: Is the plan clear and well-structured?
    \item Alignment: Does the plan align with the user's intent?
\end{enumerate}

Carefully evaluate the plan based on the criteria above. Consider whether this plan provides a solid foundation for addressing the user's request.

A plan is APPROVED (Decision: TRUE) if it:
\begin{enumerate}
    \item Directly addresses the core question or request
    \item Provides clear, actionable steps 
    \item Covers the essential aspects without being overly complex
    \item Aligns with the user's apparent intent
\end{enumerate}

A plan is REJECTED (Decision: FALSE) if it:
\begin{enumerate}
    \item Misses key aspects of the request
    \item Is too vague or abstract to act on 
    \item Contains unrealistic or impractical elements
    \item Is incomplete or poorly structured
\end{enumerate}

Respond with exactly one line in this format:
Decision: TRUE or Decision: FALSE

Please proceed with the evaluation.
Decision:

\noindent\rule{\textwidth}{1pt}

\subsection{Dataset Details} 
\label{sec:app:dataset_deats}

\begin{table}[ht]
\centering
\begin{tabular}{lc}
\toprule
\textbf{Datasets} & \textbf{Dataset size} \\
\midrule
BigCodeBench \citep{zhuo2024bigcodebench} & 50 \\
MBPP \citep{austin2021program} (Concat 2) & 50 \\
MATH500 \citep{lightman2023lets} (Concat 2) & 50 \\
Multiple Solutions & 26 \\
\bottomrule
\end{tabular}
\caption{SFT Dataset Details}
\end{table}

\begin{table}[ht]
\centering
\begin{tabular}{lc}
\toprule
\textbf{Datasets} & \textbf{Dataset size} \\
\midrule
BigCodeBench \citep{zhuo2024bigcodebench} & 500 \\
MBPP \citep{austin2021program} (Concat 2) & 50 \\
MATH500 \citep{lightman2023lets} (Concat 2) & 50 \\
Multiple Solutions & 26 \\
\bottomrule
\end{tabular}
\caption{RL Dataset Details}
\end{table}

\subsection{Training and Evaluation Datasets}
\label{app:dataset}

\noindent\textbf{SFT and RL Training Datasets.}  
We use a combination of coding and mathematical reasoning datasets to train our interleaved reasoning models.

\begin{enumerate}
    \item \textbf{BigCodeBench (BCB)} \citep{zhuo2024bigcodebench} is a benchmark of 1,140 Python programming tasks that require diverse function calls from common libraries such as \texttt{numpy} and \texttt{matplotlib}.  
    Each task includes an average of 5.6 unit tests with 99\% branch coverage.  
    For training, we split the dataset into train and test sets, and randomly sample 50 prompts from the training split to generate synthetic interleaved responses.  
    To better suit interleaved reasoning, we modify the original prompts to include not only code generation but also a brief solution outline and associated unit tests.
    
    \item \textbf{Mostly Basic Python Programs (MBPP)} \citep{austin2021program} consists of 974 crowd-sourced Python problems designed to be solvable by entry-level programmers.  
    Each problem includes a short text description and three test cases.  
    To promote multi-step reasoning, we sample multiple problems and concatenate them into a single composite prompt that requires the model to solve each sequentially.
    
    \item \textbf{MATH500} \citep{lightman2023lets} contains 500 diverse math problems spanning topics such as probability, algebra, trigonometry, and geometry.  
    Similar to MBPP, we randomly combine multiple problems into a single prompt to encourage multi-stage reasoning and plan refinement.
\end{enumerate}

\vspace{0.5em}
\noindent\textbf{Evaluation Datasets.}  
We evaluate the trained models across domains that require coding, mathematical reasoning, symbolic translation, and long-context comprehension.

\begin{itemize}
    \item \textbf{BigCodeBench (BCB)} \citep{zhuo2024bigcodebench}: challenging Python coding prompts requiring composition across multiple libraries.
    \item \textbf{Mostly Basic Python Programs (MBPP)} \citep{austin2021program}: simple Python programming tasks solvable by entry-level programmers.
    \item \textbf{MATH500} \citep{lightman2023lets}: diverse mathematical reasoning problems across algebra, probability, and geometry.
    \item \textbf{BirdSQL} \citep{li2023can}: cross-domain text-to-SQL benchmark with over 12k question–SQL pairs across 95 databases.
    \item \textbf{QuaLITY} \citep{pang2021quality}: long-document multiple-choice QA benchmark designed to test reasoning over extended contexts.
\end{itemize}


\subsection{Training Details}
\label{sec:app:train_deats}

\begin{table}[ht]
\centering
\begin{tabular}{lc}
\toprule
\textbf{Parameter} & \textbf{Value} \\
\midrule
Actor learning rate & $1 \times 10^{-6}$ \\
Critic learning rate & $1 \times 10^{-6}$ \\
Train batch size per gpu & 32 \\
Validation batch size & 256 \\
PPO mini batch size & 32 \\
PPO micro batch size & 16 \\
Critic micro batch size & 8 \\
KL coefficient & 0.001 \\
KL loss type & low variance KL \\
Max prompt length & 3096 tokens \\
Max response length & 2500 tokens \\
Sampling temperature & 0.7 \\
Number of samples per prompt & 8 \\
Stable training threshold ($\epsilon$) & 0.05 \\
Critic warmup steps & 0 \\
Evaluation frequency & 50 steps \\
Tensor model parallel size & 2 \\
\bottomrule
\end{tabular}
\end{table}

Experiments were conducted building on VERL \citep{sheng2024hybridflow}, an efficient reinforcement learning framework for post-training language models. 
We performed all experiments on 8 NVIDIA H100 GPUs with 80GB memory in a Google Cloud VM. 
We also used a consistent set of hyperparameters to ensure fair comparison between methods. 
We evaluate and save every 50 steps during training, and continue training from the last saved checkpoint if the training is interrupted (e.g., OOM). 

\subsection{Synthetic Interleaved Response Generation}
\label{sec:app:syn_gen}
We generate synthetic interleaved reasoning traces by iteratively prompting a larger model (Qwen3-32B). 
For BigCodeBench, we modify the vanilla coding prompts, to also ask for a solution outline and unit tests in addition to the code solution. 
We prompt the model to generate thought and intermediate response traces in this order: \texttt{thought} $\rightarrow$ \texttt{code outline} $\rightarrow$ \texttt{thought} $\rightarrow$ \texttt{code solution} $\rightarrow$ \texttt{thought} $\rightarrow$ \texttt{unit tests}. 
To prevent the base model from spending all its output budget on thinking, we terminate the thought generate after $N=256$ tokens, clean up the final sentence, and append a \texttt{</think>} token to mark the end of a subthought. 
Then we reprompt the model provided the previously generated thoughts as context to generate the next intermediate response. 
We provide an example of this interleaved trace at Example \ref{ex:bcb_example} below.

\tcbset{fontupper=\small, fontlower=\small}
\subsection{Training Dataset Prompt Examples}
\label{sec:app:dataset_examples}
\begin{tcolorbox}[colback=gray!5,colframe=black,title=BCB-Outline-Code-Unit]
\label{ex:bcb_example}
\textbf{Question:}

Save the provided Pandas DataFrame "df" in a pickle file with the given name, read it back for validation, and delete the intermediate file. The function should output with: 
\texttt{loaded\_df (pd.DataFrame)}: The loaded DataFrame from the specified file. 
You should write self-contained code starting with: \\

\begin{verbatim}
import pickle
import os

def task\_func(df, file_name="save.pkl"):
    # function implementation
    pass
\end{verbatim}
\vspace{0.5cm}
\textcolor{orange}
{First, outline the solution in a markdown format. Then, write the code to implement the solution. Finally, generate unit tests to test the code. Format the unit tests as a python function with a docstring. Use this exact format:} \\

\begin{verbatim}
import unittest 
from task\_func import task\_func 
class Test(unittest.TestCase): 
    def test_case_1(self): 
        # Test case 1 description 
        result = task\_func(...) 
        self.assertEqual(result, expected_value)
\end{verbatim}

\end{tcolorbox}

\begin{tcolorbox}[colback=gray!5,colframe=black,title=MATH500 Concatenate 2]
\textbf{Question:} 

Solve the following problems:     

1) Find all $p$ which satisfy both the inequalities $0\ge 54p-144$ and $0>12-20p$. Express your answer in interval notation, reducing any fractions in your answer.

2) A strictly increasing sequence of positive integers $a_1$, $a_2$, $a_3$, $\dots$ has the property that for every positive integer $k$, the subsequence $a_{2k-1}$, $a_{2k}$, $a_{2k+1}$ is geometric and the subsequence $a_{2k}$, $a_{2k+1}$, $a_{2k+2}$ is arithmetic. Suppose that $a_{13} = 2016$. Find $a_1$.

\textbf{Answer:}

\textcolor{purple}{<think>}Okay, so I need to find all values of p that satisfy both inequalities: $0 \geq 54p - 144$ and $0 > 12 - 20p$. Hmm, let me start by solving each inequality separately and then find their intersection ...\textcolor{purple}{</think>}

\textcolor{teal}{<answer>}$\left(\frac{3}{5},\frac{8}{3}\right]$\textcolor{teal}{</answer>}

\textcolor{purple}{<think>}Okay, so I need to solve this problem about a strictly increasing sequence of positive integers. The sequence has some properties related to geometric and arithmetic subsequences. ...\textcolor{purple}{</think>}

\textcolor{teal}{<answer>}504\textcolor{teal}{</answer>}

\end{tcolorbox}

\begin{tcolorbox}[colback=gray!5,colframe=black,title=MBPP Concatenate 2]
\textbf{Question:}

Solve the following coding problems: 

1) Write a function to reverse words in a given string. 

2) Write a function that matches a string that has an a followed by zero or more b's by using regex. 

You should write self-contained code for each problem starting with: 
\texttt{def task\_func(args): }

\textbf{Answer:}

\textcolor{purple}{<think>}Okay, I need to write a function that reverses the words in a given string. Let me think
about how to approach this. So, the input is a string like "Hello world", and the output should be
"world Hello"... \textcolor{purple}{</think>} \\

\textcolor{teal}{<answer>}
\begin{verbatim}
def task\_func(s):
    return ' '.join(reversed(s.split()))    
\end{verbatim}
\textcolor{teal}
{</answer>}
        
\textcolor{purple}{<think>}Okay, I need to write a function that checks if a string has an 'a' followed by zero or more
'b's using regex. Let me think about how to approach this. First, I remember that regular
expressions use certain patterns. The problem says "a followed by zero or more b's". So the 'a' must
be there, and then after it, there can be any number of 'b's, including none... \textcolor{purple}{</think>}

\textcolor{teal}{<answer>}
\begin{verbatim}
import re
def task\_func(text):
    patterns = 'ab*?'
    if re.search(patterns,  text):
            return ('Found a match!')
    else:
            return ('Not matched!')
\end{verbatim}
\textcolor{teal}{</answer>}
\end{tcolorbox}

\begin{tcolorbox}[colback=gray!5,colframe=black,title=Multiple Code Solutions]
\textbf{Question:}

Implement a function that performs basic string compression (e.g., 'aabcccccaaa' becomes
'a2b1c5a3'). Provide an iterative solution and one using itertools.

You should write self-contained code for each problem starting with: 
\texttt{def task\_func(args): }

\textbf{Answer:}

\textcolor{purple}{<think>}Okay, I need to write an iterative Python function for string compression. Let's see. The
problem says that consecutive repeated characters should be replaced by the character followed by
the count. And if the compressed string isn't shorter, return the original. First, I should think
about how to process the string. Since it's iterative, I can't use recursion...\textcolor{purple}{</think>}

\textcolor{teal}{<answer>}
\begin{verbatim}
def task\_func(s: str) -> str:
    if not s:
        return s

    compressed = []
    current_char = s[0]
    count = 1

    for char in s[1:]:
        if char == current_char:
            count += 1
        else:
            compressed.append(current_char + str(count))
            current_char = char
            count = 1

    # Append the last character and its count
    compressed.append(current_char + str(count))

    compressed_str = ''.join(compressed)
    return compressed_str if len(compressed_str) < len(s) else s
\end{verbatim}
\textcolor{teal}{</answer>}

\textcolor{purple}{<think>}Okay, I need to write a Python function called task\_func that compresses a string using
itertools. The goal is to replace consecutive repeated characters with the character followed by the
count. And if the compressed version isn't shorter, return the original. First, I remember that
itertools has groupby, which groups consecutive elements ...\textcolor{purple}{</think>}

\textcolor{teal}{<answer>}
\begin{verbatim}
from itertools import groupby

def task\_func(s: str) -> str:
    compressed = ''.join(char + str(len(list(group))) 
    for char, group in groupby(s))
    return compressed if len(compressed) < len(s) else s
\end{verbatim}
\textcolor{teal}{</answer>}

\end{tcolorbox}

\subsection{Generated Plan and Answers}
\begin{tcolorbox}[colback=gray!5,colframe=black,title=MATH500]
\textbf{Question:}

Convert the point $(0,3)$ in rectangular coordinates to polar coordinates.  Enter your answer in the form $(r,\theta),$ where $r > 0$ and $0 \le \theta < 2 \pi.$

\textbf{Plan:}

\begin{enumerate}
\item Identify the rectangular coordinates $(x, y) = (0, 3)$.
\item Use the formula $r = \sqrt{x^2 + y^2}$ to calculate the radial distance $r$.
\item Determine the angle $\theta$ using the relationship $\theta = \arctan\left(\frac{y}{x}\right)$, adjusting for the correct quadrant based on the signs of $x$ and $y$.
\item For the point $(0, 3)$, recognize that it lies on the positive $y$-axis, so $\theta = \frac{\pi}{2}$.
\item Verify that $\theta$ is within the required range $0 \leq \theta < 2\pi$.
\item Write the final polar coordinates in the form $(r, \theta)$.
\end{enumerate}

\textbf{Answer:} (3, $\frac{\pi}{2}$)

\end{tcolorbox}

\begin{tcolorbox}[colback=gray!5,colframe=black,title=BirdSQL]
\textbf{Database: debit\_card\_specializing}
\vspace{1em}

\textbf{Table: customers}
\begin{tabular}{l l}
\toprule
CustomerID & identification of the customer (integer) \\
Segment & client segment (text) \\
Currency & Currency (text) \\
\bottomrule
\end{tabular}
\vspace{1em}

\textbf{Table: gasstations}
\begin{tabular}{l l}
\toprule
GasStationID & Gas Station ID (integer) \\
ChainID & Chain ID (integer) \\
Country & No description (text) \\
Segment & chain segment (text) \\
\bottomrule
\end{tabular}
\vspace{1em}

\textbf{Table: products}
\begin{tabular}{l l}
\toprule
ProductID & Product ID (integer) \\
Description & Description (text) \\
\bottomrule
\end{tabular}
\vspace{1em}

\textbf{Table: transactions\_1k}
\begin{tabular}{l l}
\toprule
TransactionID & Transaction ID (integer) \\
Date & Date (date) \\
Time & Time (text) \\
CustomerID & Customer ID (integer) \\
CardID & Card ID (integer) \\
GasStationID & Gas Station ID (integer) \\
ProductID & Product ID (integer) \\
Amount & Amount (integer) \\
Price & Price (real) \\
\multicolumn{2}{l}{\textit{Note: total price = Amount x Price}} \\
\bottomrule
\end{tabular}
\vspace{1em}

\textbf{Table: yearmonth}
\begin{tabular}{l l}
\toprule
CustomerID & Customer ID (integer) \\
Date & Date (text) \\
Consumption & consumption (real) \\
\bottomrule
\end{tabular}

\textbf{Question:}
In 2012, who had the least consumption in LAM?

\textbf{Plan:}

\begin{enumerate}
\item Filter the \texttt{yearmonth} table for the year 2012 by checking if the `Date` starts with '2012'.
\item Join the \texttt{yearmonth} table with the \texttt{gasstations} table using the \texttt{GasStationID} to retrieve country information.
\item Filter the joined dataset to include only rows where \texttt{Country} is \texttt{LAM}.
\item Aggregate the \texttt{Consumption} values per customer using \texttt{GROUP BY CustomerID}.
\item Sort the results in ascending order of the total consumption.
\item Use \texttt{LIMIT 1} to retrieve the customer with the least consumption.
\end{enumerate}

\textbf{Answer:} 

\begin{verbatim}
SELECT y.CustomerID, SUM(y.Consumption) AS total_consumption
FROM yearmonth y
JOIN gasstations g ON y.GasStationID = g.GasStationID
WHERE LEFT(y.Date, 4) = '2012' AND g.Country = 'LAM'
GROUP BY y.CustomerID
ORDER BY total_consumption ASC
LIMIT 1;
\end{verbatim}

\end{tcolorbox}

\subsection{MATH500 Think-Answer v.s. Rewind-and-Repeat}

\begin{tcolorbox}[colback=gray!5,colframe=black,title=MATH500 Problem 153 Rewind-and-Repeat]

\textbf{Question:} The medians $AD$, $BE$, and $CF$ of triangle $ABC$ intersect at the centroid $G$.  The line through $G$ that is parallel to $BC$ intersects $AB$ and $AC$ at $M$ and $N$, respectively.  If the area of triangle $ABC$ is 144, then find the area of triangle $ENG$.

\textcolor{purple}{<think>}
...
\textcolor{purple}{</think>}

\textcolor{teal}{<answer>}
\begin{enumerate}
\item Use coordinate geometry to assign coordinates to triangle $ ABC $, simplifying calculations.
\item Determine the coordinates of the centroid $ G $ using the midpoint formula.
\item Find the equation of the line through $ G $ that is parallel to $ BC $.
\item Calculate the coordinates of points $ M $ and $ N $, where this line intersects $ AB $ and $ AC $, respectively.
\item Use the coordinates of points $ E $, $ N $, and $ G $ to compute the area of triangle $ ENG $ using the shoelace formula or determinant method.
\item Relate the area of triangle $ ENG $ to the area of triangle $ ABC $, which is given as 144, to find the numerical value.
\end{enumerate}
\textcolor{teal}{</answer>}

\textcolor{purple}{<think>}
Okay, let's start by assigning coordinates to triangle ABC. I'll place point A at $(0, 0)$, B at $(2b, 0)$, and C at $(0, 2c)$. This makes AB and AC the axes, simplifying calculations. The centroid G is the average of the coordinates, so G should be at $(\frac{2b}{3}, \frac{2c}{3})$.

Next, I need the equation of the line through G parallel to BC. The slope of BC is $\frac{2c - 0}{0 - 2b} = -\frac{c}{b}$. So the line through G with this slope is $y - \frac{2c}{3} = (-\frac{c}{b})(x - \frac{2b}{3})$. Let me rearrange this equation. Multiplying both sides by $b$ gives $b(y - \frac{2c}{3}) = -c(x - \frac{2b}{3})$. Expanding, $by - \frac{2bc}{3} = -cx + \frac{2bc}{3}$. Bringing all terms to the left: $by + cx - \frac{4bc}{3} = 0$.

Now, find point M on AB. AB is the x-axis ($y=0$). Plugging $y=0$ into the line equation: $0 + cx - \frac{4bc}{3} = 0 \Rightarrow cx = \frac{4bc}{3} \Rightarrow x = \frac{4b}{3}$. So M is at $(\frac{4b}{3}, 0)$.

Similarly, find point N on AC. AC is the y-axis ($x=0$). Plugging $x=0$ into the line equation: $by + 0 - \frac{4bc}{3} = 0 \Rightarrow by = \frac{4bc}{3} \Rightarrow y = \frac{4c}{3}$. So N is at $(0, \frac{4c}{3})$.

Now, need coordinates for E. Wait, E is the midpoint of AC because BE is a median. Since A is $(0,0)$ and C is $(0,2c)$, the midpoint E is at $(0, c)$.

So triangle ENG has points $E(0, c)$, $N(0, \frac{4c}{3})$, and $G(\frac{2b}{3}, \frac{2c}{3})$. Let's compute its area. Using the shoelace formula:

Area = $\frac{1}{2} | (x_E(y_N - y_G) + x_N(y_G - y_E) + x_G(y_E - y_N)) |$. Plugging in:

$x_E = 0, y_E = c; x_N = 0, y_N = \frac{4c}{3}; x_G = \frac{2b}{3}, y_G = \frac{2c}{3}$.

Calculating each term:

First term: $0 \cdot (\frac{4c}{3} - \frac{2c}{3}) = 0 \cdot (\frac{2c}{3}) = 0$.

Second term: $0 \cdot (\frac{2c}{3} - c) = 0 \cdot (-\frac{c}{3}) = 0$.

Third term: $(\frac{2b}{3}) \cdot (c - \frac{4c}{3}) = (\frac{2b}{3}) \cdot (-\frac{c}{3}) = -\frac{2bc}{9}$.

Taking absolute value and multiplying by $\frac{1}{2}$: $\frac{1}{2} \cdot | -\frac{2bc}{9} | = \frac{1}{2} \cdot \frac{2bc}{9} = \frac{bc}{9}$.

So the area of ENG is $\frac{bc}{9}$.

Now, relate this to the area of ABC. The area of ABC is 144. Since ABC is a right triangle with legs $2b$ and $2c$, its area is $(\frac{1}{2}) \cdot 2b \cdot 2c = 2bc$. So $2bc = 144 \Rightarrow bc = 72$.

Therefore, the area of ENG is $\frac{72}{9} = 8$.

Wait, that seems straightforward. Let me check each step again. The coordinates assignment, centroid calculation, line equation, points M and N, and area computation. All steps seem correct. The key was setting up the coordinate system and using the shoelace formula. The answer should be 8.
\textcolor{purple}{</think>}

\textcolor{teal}{<answer>}8
\textcolor{teal}{</answer>}

\end{tcolorbox}

\label{sec:app:math500_case_study}
\begin{tcolorbox}[colback=gray!5,colframe=black,title=MATH500 Problem 153 Think-Answer]

\textbf{Question:} The medians $AD$, $BE$, and $CF$ of triangle $ABC$ intersect at the centroid $G$.  The line through $G$ that is parallel to $BC$ intersects $AB$ and $AC$ at $M$ and $N$, respectively.  If the area of triangle $ABC$ is 144, then find the area of triangle $ENG$.

\textcolor{purple}{<think>}
Okay, so I need to find the area of triangle $ENG$ in triangle $ABC$, where the medians $AD$, $BE$, and $CF$ intersect at centroid $G$. There's a line through $G$ parallel to $BC$ that intersects $AB$ at $M$ and $AC$ at $N$. The area of $ABC$ is 144. Hmm, let me try to visualize this first.

First, let me recall some properties of centroids. The centroid divides each median into a ratio of 2:1, with the longer part being closer to the vertex. So, for example, $AG:GD = 2:1$, $BG:GE = 2:1$, and $CG:GF = 2:1$.

Now, there's a line through $G$ parallel to $BC$. Since $BC$ is the base of the triangle, this line is going to be somewhere inside the triangle. It intersects $AB$ at $M$ and $AC$ at $N$. So, $MN$ is parallel to $BC$ and passes through $G$. I need to figure out where exactly $M$ and $N$ are located on $AB$ and $AC$.

Since $MN$ is parallel to $BC$, triangle $AMN$ is similar to triangle $ABC$ by the Basic Proportionality Theorem or Thales' theorem. Wait, but $MN$ isn't necessarily a midline unless it's halfway. However, since it passes through the centroid, which divides the medians in a 2:1 ratio, maybe I can find the ratio of similarity.

Let me think. Let me consider coordinates. Maybe coordinate geometry can help here. Let me assign coordinates to triangle $ABC$. Let me place point $A$ at $(0, 0)$, $B$ at $(2b, 0)$, and $C$ at $(2c, 2d)$. Wait, but maybe it's better to use coordinates that make calculations easier. Since the centroid divides the medians in 2:1, maybe choosing coordinates where the centroid is at $(0, 0)$? Wait, but maybe that complicates things. Alternatively, let me use standard coordinates.

Alternatively, let me take coordinates such that $BC$ is horizontal for simplicity. Wait, but since $MN$ is parallel to $BC$, maybe it's better to have $BC$ horizontal. Let me try.

Let me set coordinates:

Let me let $BC$ be on the x-axis for simplicity. Let me set point $B$ at $(0, 0)$, point $C$ at $(c, 0)$, and point $A$ somewhere in the plane, say $(a, b)$. Then, the centroid $G$ is the average of the coordinates of $A$, $B$, and $C$. So, $G_x = (a + 0 + c)/3$, $G_y = (b + 0 + 0)/3 = b/3$.

Now, the line through $G$ parallel to $BC$. Since $BC$ is horizontal (from $(0,0)$ to $(c,0)$), it's horizontal, so the line through $G$ parallel to $BC$ is also horizontal. Therefore, it has equation $y = G_y = b/3$. This line intersects $AB$ and $AC$ at points $M$ and $N$.

So, I need to find points $M$ on $AB$ and $N$ on $AC$ such that their y-coordinate is $b/3$.

Let me parametrize $AB$ and $AC$.

First, equation of $AB$: from $A(a, b)$ to $B(0, 0)$. The parametric equations can be written as $x = a - a \cdot t, y = b - b \cdot t$, where $t$ ranges from 0 to 1. Wait, or maybe better to use a parameter $s$.

Alternatively, let me find the equation of $AB$. The slope of $AB$ is $\frac{0 - b}{0 - a} = \frac{b}{a}$. So equation is $y = (\frac{b}{a})x$.

Similarly, equation of $AC$: from $A(a, b)$ to $C(c, 0)$. The slope is $\frac{0 - b}{c - a} = -\frac{b}{c - a}$. Equation is $y - b = [-\frac{b}{c - a}](x - a)$.

Now, the line $y = b/3$ intersects $AB$ and $AC$. Let me find coordinates of $M$ and $N$.

For point $M$ on $AB$: $y = (\frac{b}{a})x = \frac{b}{3}$. Therefore, $x = (\frac{b}{3}) \cdot (\frac{a}{b}) = \frac{a}{3}$. Therefore, coordinates of $M$ are $(\frac{a}{3}, \frac{b}{3})$.

Similarly, for point $N$ on $AC$: Let me solve $y = \frac{b}{3}$ with equation of $AC$.

Equation of $AC$: $y = [-\frac{b}{c - a}](x - a) + b$.

Set $y = b/3$:

$\frac{b}{3} = [-\frac{b}{c - a}](x - a) + b$

...

Wait no. Wait no. Wait no. Wait no...
\textcolor{purple}{</think>}

\end{tcolorbox}

\end{appendices}

\end{document}